\documentclass[]{bytedance_seed}

\usepackage[toc,page,header]{appendix}

\usepackage{minitoc}

\usepackage{enumitem}
\usepackage{pifont}   

\usepackage{xspace}
\usepackage{xcolor}
\usepackage{colortbl}
\usepackage{url}
\usepackage{mathtools}
\usepackage{amsmath}
\usepackage{amsthm}
\usepackage{amssymb}
\usepackage{graphicx}
\usepackage{wrapfig}
\usepackage{subcaption}

\usepackage{array}

\usepackage{bbm}
\usepackage{bm}
\usepackage{multirow}
\usepackage{booktabs}
\usepackage{xspace}
\usepackage{tcolorbox}
\usepackage{lipsum} 
\usepackage{tabularx}
\usepackage{algorithm}  
\usepackage[endLComment=]{algpseudocodex}

\definecolor{myDeepYellow}{rgb}{0.9412, 0.6902, 0.302}
\definecolor{myYellow}{rgb}{0.9765, 0.8824, 0.7255}
\definecolor{myBlue}{rgb}{0.6353, 0.7686, 0.8627}
\definecolor{lightgray}{gray}{0.93}
\definecolor{bestrow}{RGB}{235,245,255}

\def\docommandbetter#1 {\colorbox{myBlue!80}{#1} \let\next\argii}
\def\argii{\let\next\docommandbetter}

\makeatletter
\newcommand{\smartperiod}{\@ifnextchar.{}{.\@\xspace}}
\newcommand{\smartcomma}{\@ifnextchar.{}{,}\xspace}
\makeatother

\newcommand{\geminitwo}{\textsc{Gemini-2.5-Pro}\xspace}
\newcommand{\arflow}{\textsc{AR-Flow}\xspace}
\newcommand{\method}{\textsc{FlowPortrait}\xspace}

\newcommand{\pos}[1]{\textcolor{blue!60!black}{#1}}
\newcommand{\negval}[1]{\textcolor{orange!80!black}{#1}}

\title{FlowPortrait: Reinforcement Learning for Audio-Driven Portrait Video Generation}

\author[\spadesuit\heartsuit *,\dagger]{Weiting Tan}
\author[\spadesuit, \dagger]{\,\,\,Andy T. Liu}
\author[\spadesuit]{\,\,\,Ming Tu}
\author[\spadesuit]{\,\,\,Xinghua Qu}
\author[\heartsuit]{\\Philipp Koehn}
\author[\spadesuit]{\,\,\,\,Lu Lu}

\affiliation[\spadesuit]{ByteDance Seed}
\affiliation[\heartsuit]{Johns Hopkins University}

\contribution[*]{Work done at ByteDance Seed}
\contribution[\dagger]{Corresponding authors}

\abstract{
Generating realistic talking-head videos remains challenging due to persistent issues such as imperfect lip synchronization, unnatural motion, and evaluation metrics that correlate poorly with human perception. We propose \method, a reinforcement-learning framework for audio-driven portrait animation built on a multimodal backbone for autoregressive audio-to-video generation. \method introduces a human-aligned evaluation system based on Multimodal Large Language Models (MLLMs) to assess lip-sync accuracy, expressiveness, and motion quality. These signals are combined with perceptual and temporal consistency regularizers to form a stable composite reward, which is used to post-train the generator via Group Relative Policy Optimization (GRPO). Extensive experiments, including both automatic evaluations and human preference studies, demonstrate that \method consistently produces higher-quality talking-head videos, highlighting the effectiveness of reinforcement learning for portrait animation.
}

\date{\today}
\correspondence{Weiting Tan at \email{wtan12@jhu.edu}, Andy T. Liu at \email{tingwei.liu@bytedance.com}}

\begin{document}
\maketitle
\begingroup
\renewcommand{\thefootnote}{}
\footnotetext{This work is for research purposes only and is not integrated into any ByteDance products.}
\endgroup

\section{Introduction}\label{sec::intro}

Portrait animation—generating realistic, expressive talking-head videos from a single image and an audio clip—has advanced rapidly, driven by applications in virtual avatars, video conferencing, and digital entertainment. Earlier approaches relied on explicit facial landmarks \cite{gao2023highfidelityfreelycontrollabletalking, zhang2023metaportraitidentitypreservingtalkinghead}, whereas recent systems increasingly adopt end-to-end diffusion frameworks. Architectures such as U-Net \cite{unet} and Diffusion Transformers (DiT) \citep[DiT]{dit}, together with improvements in contextual modeling, emotion conditioning, and motion-decoupled control modules \cite{sonic,memo,echomimic}, have enhanced visual fidelity and expressiveness.

Despite this progress, two major challenges remain. First, most prior portrait animation models rely on training DiT-style backbones from scratch and conditioning them on audio features, which limits their ability to exploit rich cross-modal priors. In contrast, recent advances in Multimodal Large Language Models (MLLMs) and video-generation backbones~\cite{bagel,stable_video_diffusion,janus_flow,intern_video,cog_video} have demonstrated the effectiveness of large-scale pretraining for multimodal reasoning and generation. Motivated by these developments, we propose \method, an audio-driven portrait animation framework built on BAGEL~\cite{bagel}, a unified MLLM based on the Autoregressive Rectified Flow (\arflow) architecture. By casting audio-to-video generation as an autoregressive process within a pretrained multimodal model, \method effectively transfers large-scale cross-modal knowledge to portrait animation while naturally supporting future extensions such as multi-hop audio–video generation.

Second, evaluation metrics remain a major bottleneck. Widely used measures such as PSNR and SSIM emphasize pixel-level correspondence and fail to capture perceptual factors central to portrait animation, including lip-sync accuracy, emotional expressiveness, and motion naturalness. While FVD \cite{fvd} and SyncNet-based scores \citep[LSE-C/D]{syncnet} address some of these dimensions, they remain imperfect proxies for human judgment and often overlook fine-grained semantic or temporal artifacts. To address this gap, \method introduces an MLLM-based evaluation framework that decomposes assessment into three specialized agents—lip-sync, expressiveness, and motion—whose aggregated judgments provide a more human-aligned and diagnostically informative evaluation signal.

\begin{figure}[t]
    \centering
    \includegraphics[width=0.95\linewidth]{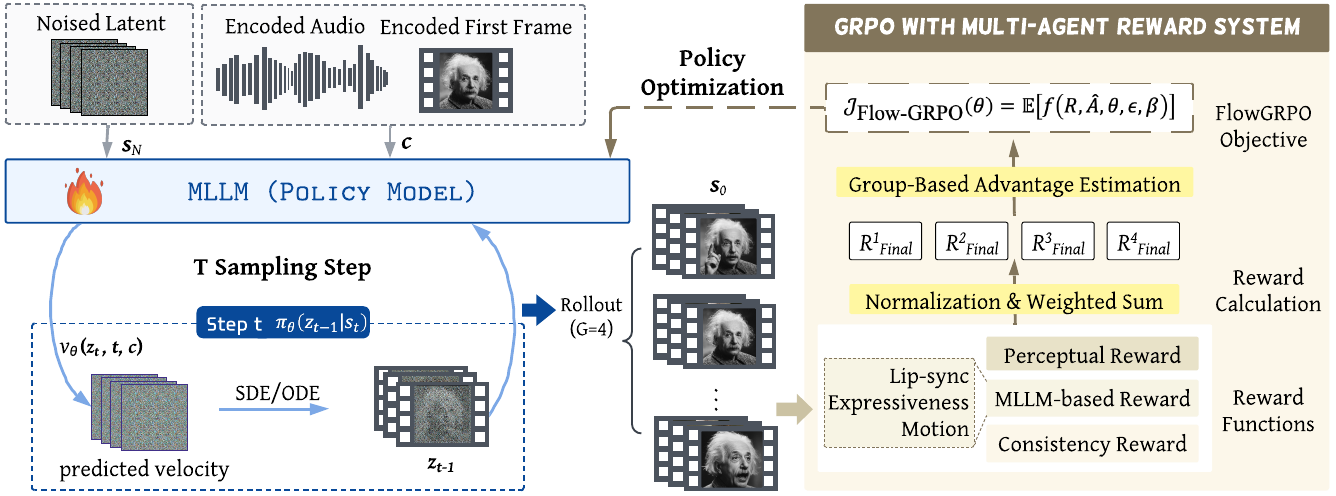}
    \caption{Overview of \method. We build upon a pretrained MLLM to enable audio-to-video generation for portrait animation. Using an improved Flow-GRPO objective, we post-train the \arflow generator with a composite reward that integrates MLLM-based evaluations with perceptual and temporal consistency terms, leading to consistent improvements in generation quality.}
    \label{fig:intro}
\end{figure}

Finally, \method unifies modeling and evaluation through reinforcement learning. Using an improved Flow-GRPO objective \cite{flow_grpo, tempflow_grpo, cps_sampling}, we post-train the \arflow generator directly against a composite reward that combines MLLM-based evaluations with perceptual and temporal consistency rewards. This design mitigates reward hacking and enables stable, long-horizon improvements in generation quality. In summary, our contributions are:

\begin{itemize}[leftmargin=2em, itemsep=2pt, topsep=2pt]
\item \method, a speech-driven portrait animation framework built on a pretrained \arflow-based MLLM, enabling autoregressive audio-to-video generation with strong multimodal priors. \vspace{-5pt}
\item A MLLM-based evaluation framework that separately scores lip-sync, expressiveness, and motion smoothness, yielding a more human-aligned and interpretable assessment of portrait animation quality. \vspace{-6pt}
\item A reinforcement learning pipeline based on Flow-GRPO using a composite reward system, allowing FlowPortrait to be post-trained for improved quality and temporal coherence. \vspace{-6pt}
\end{itemize}

\section{Automatic Evaluation for Portrait Annimation}\label{sec::prelim}

\subsection{Existing Evaluation Metrics}
Recent portrait animation methods \cite{sonic, memo, echomimic} predominantly rely on hand-crafted evaluation rubrics derived from low-level image or video features. For instance, FID \cite{fid} and FVD \cite{fvd} measures distributional discrepancies between real and generated image/video representations, while LSE-C/D \cite{syncnet} assesses lip-synchronization accuracy using a pre-trained lip-sync model. Other commonly adopted metrics include pixel-level similarity measures such as PSNR (Peak Signal-to-Noise-Ratio) and SSIM (Structural Similarity Index Measure). More recently, \cite{quignon2025thevalevaluationframeworktalking} proposes a composite evaluation framework that combines multiple criteria related to visual quality, naturalness, and synchronization.

Despite their popularity, these metrics have been shown to correlate poorly with human perceptual judgments, particularly for generative tasks involving complex motion, semantics, and realism. Pixel-based measures such as PSNR and SSIM are known to favor low-level fidelity while failing to reflect semantic similarity as perceived by humans \cite{wang2004ssim, zhang2018unreasonableeffectivenessdeepfeatures}. Although FVD exhibits stronger alignment with human judgment than frame-level metrics, it remains sensitive to feature representations, sample size, and distributional assumptions, and often fails to capture fine-grained temporal artifacts and motion naturalness \cite{unterthiner2019accurategenerativemodelsvideo, blau2018perception}. Moreover, these metrics require access to paired ground-truth videos, which is frequently unavailable in real-world portrait animation scenarios, further limiting their practical reliability.

\subsection{Automatic Evaluation via Multimodal LLMs}\label{sec::mllm_evaluation}

Due to the limitations of rubric-based evaluation metrics, we explore an alternative evaluation framework based on \emph{multimodal large language models (MLLMs)} for portrait animation assessment. We investigate two complementary paradigms: \textbf{score-based evaluation} and \textbf{comparison-based evaluation}.

In \textbf{score-based evaluation}, an MLLM assigns a numerical score in the range $[1,5]$ to each video. For any video pair, the final preference is determined by comparing their predicted scores. In \textbf{comparison-based evaluation}, the model is directly prompted to compare two videos and select the better one. To mitigate positional bias~\cite{shi2025judgingjudgessystematicstudy}, we randomize the order of the two videos in each prompt.

We adopt \geminitwo as the backbone for all evaluation systems\footnote{\geminitwo is used solely for evaluation purposes and is not involved in any training process.}, motivated by its strong video understanding and multimodal reasoning capabilities. In preliminary experiments, we also evaluated \textsc{Qwen3-Omni} but found its outputs to be highly saturated, with scores almost always concentrated at 4 or 5. As a result, it failed to meaningfully differentiate the quality of outputs from different generators (e.g., Sonic and EchoMimics yielding nearly identical dataset-level scores despite clear perceptual differences in human evaluations). Consequently, all reported results are based on \geminitwo. We explore the following variations for each evaluation paradigm, with their detailed prompt templates provided in \cref{app::evaluation_prompts}.

We consider three score-based variants that differ in whether evaluation is performed holistically or decomposed across aspects, and whether judgments are produced by a single or multiple MLLMs.

\begin{itemize}[leftmargin=2em, itemsep=2pt, topsep=2pt]
    \item \textbf{SAS-SA} (Single-Aspect, Single-Agent): A single MLLM provides one holistic score in $[1,5]$ following the human annotation rubric. \vspace{-6pt}
    \item \textbf{MAS-SA} (Multi-Aspect, Single-Agent): A single MLLM predicts separate scores for lip-sync, expressiveness, and motion, which are then averaged to form the final score. \vspace{-6pt}
    \item \textbf{MAS-MA} (Multi-Aspect, Multi-Agent): Each aspect is evaluated by a dedicated MLLM (all instantiated with \geminitwo), and their outputs are aggregated into the final judgment.
\end{itemize}

Similarly, for comparison-based evaluator, we consider following variants: 
\begin{itemize}[leftmargin=2em, itemsep=2pt, topsep=2pt]
    \item \textbf{Direct-Comp}: Performs a holistic, forced-choice comparison between two videos using a single MLLM. \vspace{-6pt}
    \item \textbf{ICL-Comp}: Enhances the comparison prompt with three in-context examples—one per evaluation aspect—to guide more structured reasoning within a single MLLM. \vspace{-6pt}
    \item \textbf{MA-Comp} (Multi-Agent Comparison): Decomposes evaluation across aspects, where three specialized MLLMs independently assess lip-sync, expressiveness, and motion, and a fourth MLLM aggregates their decisions into a final preference.
\end{itemize}

\subsection{Automatic Evaluation's Alignment with Human Judgment}\label{sec::human_alignment}

To evaluate our proposed evaluation systems, we first collect human judgment data from an in-house talking-head video dataset. We randomly sample 940 portrait videos that are at least 5 seconds long and have high resolution (with a detected head region of at least $512\times512$; see \cref{app::preprocessing} for details on preprocessing) to form our test split. For each sample, we generate portrait animations from the input audio and the first ground-truth frame using three open-sourced methods: Sonic \cite{sonic}, Memo \cite{memo}, and Echomimic \cite{echomimic}. Together with the ground-truth video, this results in four versions per sample. We ask three human annotators to score and rank these videos according to lip-sync quality, facial expressiveness, and motion smoothness, following the detailed guidelines in \cref{app::human_annotation}.

After collecting both scores and rankings, we construct a pairwise dataset to measure the alignment between human judgments and automatic evaluation metrics. Specifically, for each sample, we form all possible pairs among the four videos ($\binom{4}{2}=6$ pairs) and retain only those pairs for which all three annotators unanimously agree on the preferred video. The agreed-upon video is labeled as the winner. This process yields 4,501 human-labeled preference pairs, corresponding to 80.7\% of all possible pairs. We then apply the evaluation systems described in \cref{sec::mllm_evaluation} to predict the preferred video for each pair and compute accuracy as the percentage of pairs for which the predicted winner matches the human consensus.

\begin{wraptable}{r}{0.60\textwidth}
\vspace{-1em}
\centering
\small
\setlength{\tabcolsep}{6pt}
\renewcommand{\arraystretch}{1.15}

\begin{tabular}{l l c c c}
\toprule
\textbf{Base Model} & \textbf{Method}
& \textbf{Acc}
& \textbf{Acc$_{\text{nt}}$}
& \textbf{Cov. (\%)} \\
\midrule

\rowcolor{gray!6}
\multicolumn{5}{l}{\textbf{Feature / Pixel Models}} \\
Inception-V3 \cite{inceptionv3} & FID   & 27.5 & -- & -- \\
SyncNet \cite{syncnet}          & LSE-C & 53.4 & -- & -- \\
SyncNet                          & LSE-D & 51.6 & -- & -- \\
Pixel-based                      & SSIM  & 61.2 & -- & -- \\
\midrule

\rowcolor{gray!6}
\multicolumn{5}{l}{\textbf{Comparison Models}} \\
 & Direct-Comp & 64.3 & -- & -- \\
Gemini~2.5       & ICL-Comp    & 64.3 & -- & -- \\
\rowcolor{bestrow}
& \textbf{MA-Comp} & \textbf{66.6} & \textbf{--} & \textbf{--} \\
\midrule

\rowcolor{gray!6}
\multicolumn{5}{l}{\textbf{Score Models}} \\
 & SAS-SA   & 58.3 & 67.2 & 46.8 \\
Gemini~2.5         & MAS-SA   & 59.7 & 65.0 & 64.9 \\
\rowcolor{bestrow}
 & \textbf{MAS-MA} & \textbf{65.6} & \textbf{69.3} & \textbf{81.1} \\
\bottomrule
\end{tabular}

\caption{
Evaluation against human preferences.
For score-based methods, \textbf{Acc} assigns half credit to ties.
\textbf{Cov.} is the proportion of non-tied predictions, and
Acc$_{\text{nt}}$ is accuracy among non-tied samples.
}
\vspace{-1.5em}
\label{tab::human_correlation}
\end{wraptable}

We choose FID, FVD, LSE-C, LSE-D, and SSIM as representative traditional metrics for comparison.
For FID, we use Inception-V3 \cite{inceptionv3} to extract frame-level features, average them across frames to obtain video-level representations. We compute the FVD metric using a 3D ResNet-18 \cite{he2015deepresiduallearningimage} pretrained on Kinetics-400 \cite{kay2017kineticshumanactionvideo} with standard Kinetics normalization, treating video-level features as a Gaussian distribution and reporting the Fréchet distance to the ground-truth distribution. For LSE-C and LSE-D, we follow the official implementation \cite{syncnet} to compute lip-sync confidence and distance scores. SSIM is computed at the frame level with Scikit-Image's implementation \footnote{~Available at \url{https://scikit-image.org/}} and averaged across all frames. As shown in \cref{tab::human_correlation}, prior metrics LSE-C and LSE-D achieve only 53.4\% and 51.6\% accuracy, respectively, \emph{indicating weak alignment with human preferences} on portrait animation quality. Similarly, using SSIM and FID also yields poor alignment.

This observation is consistent with prior findings that such metrics correlate poorly with human perception \cite{quignon2025thevalevaluationframeworktalking}. Among MLLM-based approaches, multi-agent systems (MA-Comp and MAS-MA) achieve the strongest performance, suggesting that decomposing evaluation across specialized agents for different aspects leads to more accurate judgments. Overall, comparison-based and score-based methods obtain comparable accuracy. We therefore adopt MAS-MA as our default evaluation method throughout the paper, as it is more cost-effective (comparison-based methods require encoding two videos per prompt) and better suited for RL post-training scenarios, where each sampled video can be scored directly.

Moreover, \cref{tab::generator_scores} reports the scores assigned to different video generators by traditional automatic metrics, MLLM-based evaluators, and human annotators. We observe that MLLM-based evaluators (SAS-SA, MAS-SA, and MAS-MA) exhibit substantially better alignment with human judgments, consistently producing the ranking Original $>$ Sonic $>$ Memo $>$ Echomimics. In contrast, conventional metrics such as SSIM, LSE-C and LSE-D fail to accurately reflect this ordering. These results further highlight the effectiveness of MLLM-based evaluation systems in capturing human-perceived video quality.

\begin{table}[t]
\centering
\small
\setlength{\tabcolsep}{5pt}
\begin{tabular}{lccccc|ccc|c}
\toprule
\textbf{Model} 
& \multicolumn{9}{c}{\textbf{Evaluation System}} \\
\cmidrule(lr){2-10}
& \textbf{FVD}$_{\downarrow}$
& \textbf{FID}$_{\downarrow}$
& \textbf{LSE-D}$_{\downarrow}$ 
& \textbf{LSE-C}$_{\uparrow}$ 
& \textbf{SSIM}$_{\uparrow}$ 
& \textbf{SAS-SA}$_{\uparrow}$ 
& \textbf{MAS-SA}$_{\uparrow}$ 
& \textbf{MAS-MA}$_{\uparrow}$ 
& \textbf{Human}$_{\uparrow}$ \\
\midrule
Original   
& --    & --    & --    & --    & --     & 3.32 & 3.76 & 3.93 & 4.77 \\
Sonic      
& 62.25 & 44.14 & 12.05 & 2.91  & 0.633  & 2.94 & 3.43 & 3.58 & 3.56 \\
Memo       
& 74.94 & 41.12 & 13.65 & 0.92  & 0.668  & 2.72 & 3.31 & 3.37 & 3.21 \\
Echomimics 
& 97.06 & 60.15 & 13.18 & 1.51  & 0.573  & 2.55 & 3.16 & 3.26 & 2.54 \\
\bottomrule
\end{tabular}

\caption{
Scores assigned to different generation models by automated metrics, MLLM-based evaluators, and human annotators. Arrows indicate whether higher ($\uparrow$) or lower ($\downarrow$) values are better. Vertical separators partition evaluation systems into (i) automated metrics, (ii) MLLM-based evaluators, and (iii) human judgments.
}
\vspace{-1.5em}
\label{tab::generator_scores}
\end{table}

\subsection{Deep Dive into Multi-Aspect, Multi-Agent Evaluation}

\begin{figure}[h]
    \centering
    \includegraphics[width=0.9\linewidth]{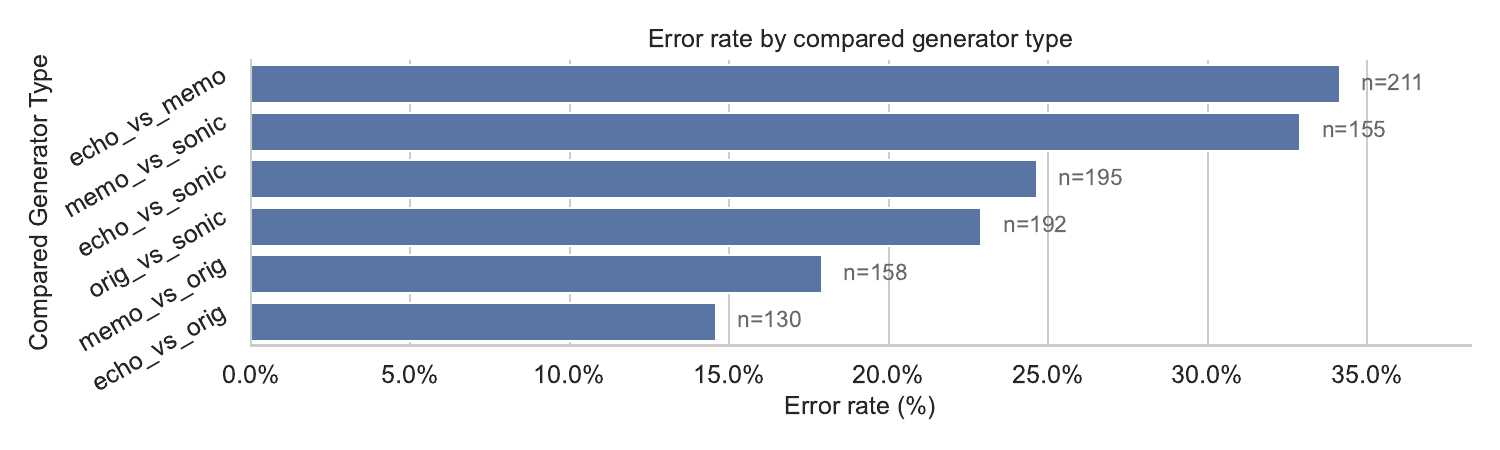}
    \vspace{-1.5em}
    \caption{Error rates of different generator types. Each bar represents the error rate for a specific generator type, with annotations indicating the number of error cases.}
    \label{fig:agent_score_analysis}
\end{figure}

Having established the effectiveness of MAS--MA, we further analyze its per-aspect behavior and score distributions.

We first examine the error rate of MAS--MA in \cref{fig:agent_score_analysis}. Most disagreements with human judgments occur when comparing generators of similar quality, such as \textit{Echomimics} vs.\ \textit{Memo} and \textit{Memo} vs.\ \textit{Sonic}. As shown in \cref{tab::generator_scores}, both humans and MAS--MA rank the generators consistently as \textit{Original} $>$ \textit{Sonic} $>$ \textit{Memo} $>$ \textit{Echomimics}, confirming that the system is reliable when the quality gap is sufficiently large.

\begin{figure}[t]
    \centering
    \includegraphics[width=0.9\linewidth]{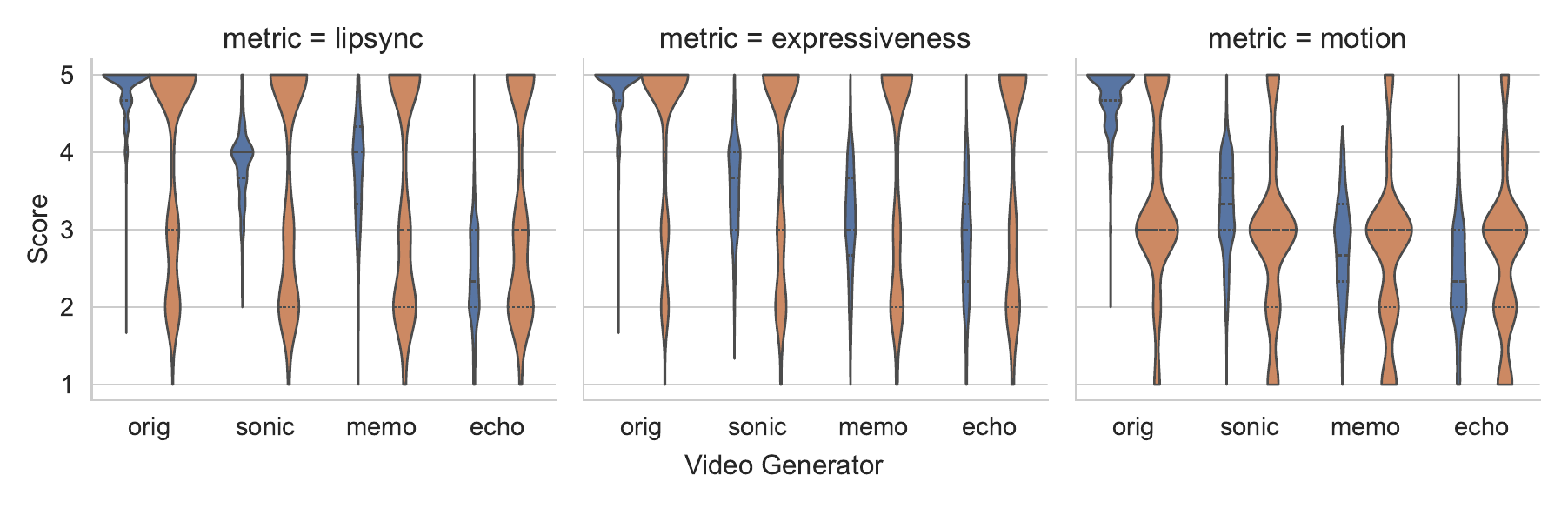}
    \vspace{-1.5em}
    \caption{Score distribution histograms for the three evaluation aspects: lip-sync quality, facial expressiveness, and motion smoothness. Each histogram illustrates the frequency of scores assigned by Human (blue) and MAS-MA evaluation system (orange) across the range of possible scores (1 to 5).}
    \vspace{-1em}
    \label{fig:score_distribution}
\end{figure}

We next compare the score distributions of humans and MAS--MA in \cref{fig:score_distribution}. MAS--MA produces less calibrated distributions compared to human ratings: lip-sync and expressiveness scores tend toward a binary pattern (very high or very low), while motion smoothness scores cluster in the mid-range with occasional extreme lows.

These discrepancies highlight remaining gaps between automatic and human evaluation. Achieving more human-like score distributions will likely require larger, higher-quality annotated datasets for reward modeling—a direction we leave for future work.


\section{Training Methodology}\label{sec::approach}

\textbf{FlowPortrait} is trained using a two-stage pipeline for speech-driven portrait animation. In the first stage, we fine-tune our MLLM backbone--BAGEL--to generate video frames conditioned on speech. In the second stage, we perform post-training with a reinforcement learning framework, Flow-GRPO \cite{flow_grpo}, to further improve generation quality using our proposed reward system.

\subsection{Autoregressive Rectified Flow Model}
\label{sec::rectified_diffusion_transformer}

Given an input audio clip $\bm{x}$ and a reference image $\bm{i}$, our goal is to generate a sequence of video frames $\bm{v} = \{\bm{v}_1, \cdots, \bm{v}_T\}$ that are temporally coherent and synchronized with the audio to create a portrait animation.

First, we encode the ground-truth video frames $\bm{v}$ into a latent sequence $\bm{z}_{0} = \{\bm{z}_{0}^1, \cdots, \bm{z}_{0}^L\}$ using a pre-trained VAE encoder \citep{wan2025wanopenadvancedlargescale}. The latent sequence length $L = T/R$ is shorter than the frame count $T$ due to the VAE's temporal reduction factor $R$.

We adopt the \textbf{Rectified Flow} \cite{rectified_flow} framework for generation. This model defines a linear interpolation between the true data sample $\bm{z}_0$ (the latent sequence) and a noise sample $\bm{z}_1 \sim \mathcal{N}(0, \mathbf{I})$ of the same shape. The noised latent $\bm{z}_t$ at continuous time $t \in [0, 1]$ is defined as:
\begin{equation}
    \bm{z}_t = (1 - t) \bm{z}_0 + t \bm{z}_1
\end{equation}
Our model follows the Mixture of Transformers (MoT) structure \citep{mixture_of_transformers} from BAGEL \cite{bagel}. Conditional information $\bm{c}$, which includes audio features and reference image ViT tokens, is processed by an "understanding expert" transformer. A separate "generation expert" transformer, $v_\theta$, is trained to predict the velocity field $\bm v_\theta(\bm{z}_t, t, \bm{c})$ given the noised latent $\bm{z}_t$, timestep $t$, and conditions $\bm{c}$. The model is trained by minimizing the conditional Flow Matching loss:
\begin{equation}
    \mathcal{L}(\theta) = \mathbb{E}_{t, \bm{z}_0, \bm{z}_1, \bm{c}} \left[ \Vert (\bm{z}_1 - \bm{z}_0) - \bm v_\theta(\bm{z}_t, t, \bm{c}) \Vert^2 \right]
\end{equation}
This objective trains $v_\theta$ to learn the constant target velocity $\bm{z}_1 - \bm{z}_0$ required to move from noise $\bm{z}_1$ to data $\bm{z}_0$. During generation (the reverse process from $t=1$ to $t=0$), the trained velocity model $v_\theta$ can be used to predict the data ($\hat{\bm{z}}_0$) and noise ($\hat{\bm{z}}_1$) components from any intermediate state $\bm{z}_t$:
\begin{equation}
    \hat{\bm{z}}_0(\bm{z}_t, t, \bm{c}) = \bm{z}_t - t \cdot \bm v_\theta(\bm{z}_t, t, \bm{c})
\end{equation}
\begin{equation}
    \hat{\bm{z}}_1(\bm{z}_t, t, \bm{c}) = \bm{z}_t + (1-t) \cdot \bm v_\theta(\bm{z}_t, t, \bm{c})
\end{equation}
To control information flow, a specific attention mask is applied:
\begin{itemize}
    \item \textbf{Within blocks:} Full self-attention is used within the audio, image, and video latent blocks, respectively.
    \item \textbf{Across blocks:} A causal-style mask is applied. The audio block attends only to itself. The reference image block attends to audio and image tokens. The video latent block (the generation expert) attends to all inputs: audio, image, and its own latents.
\end{itemize}

\subsection{Reinforcement Learning with Flow-GRPO}
\label{sec::flow_grpo}

To further enhance video generation quality, we incorporate reinforcement learning using Flow-GRPO~\cite{flow_grpo, tempflow_grpo}. This allows the model to optimize directly for our customized reward function rather than merely imitating ground-truth videos through supervised learning. We begin by formulating the generative process as a Markov Decision Process (MDP) discretized into $N$ steps (from $t = N$ down to $0$). At each step $t$, the policy $\pi_\theta(\bm{z}_{t-1} \mid \bm{s}_t)$, parameterized by $v_\theta$, predicts the next latent $\bm{z}_{t-1}$ based on the state $\bm{s}_t \triangleq (\bm{c}, t, \bm{z}_t)$, starting from an initial noise state $\bm{s}_N \sim \mathcal{N}(0, \mathbf{I})$. A sparse reward $R(\bm{z}_0, \bm{c})$ is assigned only at the final step, scoring the generated video according to our evaluation system.

\textbf{Group-Based Advantage Estimation}
Given this MDP formulation, we adopt GRPO~\cite{grpo} to estimate policy advantages using group-relative normalization. For each condition $\bm{c}$, we sample a group of $G$ trajectories $\{\bm{z}_N^i, \ldots, \bm{z}_0^i\}_{i=1}^G$ using the old policy $\pi_{\theta_{\text{old}}}$. The advantage of trajectory $i$ is computed as
\begin{equation}
    \hat{A}_i = \frac{R(\bm{z}_0^i, \bm{c}) - \text{mean}(\{R(\bm{z}_0^j, \bm{c})\}_{j=1}^G)}
    {\text{std}(\{R(\bm{z}_0^j, \bm{c})\}_{j=1}^G)}.
\end{equation}

\textbf{Flow-GRPO Objective}
Flow-GRPO extends GRPO to Flow-based generative models, optimizing the objective:
\begin{equation}\label{eq::flow_grpo}
    J_{\text{Flow-GRPO}}(\theta) = \mathbb{E}_{\bm{c} \sim \mathcal{C}, \{\bm{z}_{N\dots 0}^i\}_{i=1}^G \sim \pi_{\theta_{\text{old}}}(\cdot|\bm{c}) } \left[ f(R, \hat{A}, \theta, \varepsilon, \beta) \right]
\end{equation}
where $f$ is defined as:
\begin{equation}
    f(\cdot) = \frac{1}{G} \sum_{i=1}^G \frac{1}{N} \sum_{t=1}^{N} \left[ \min\left(r_t^i(\theta)\hat{A}_i, \text{clip}(r_t^i(\theta), 1-\varepsilon, 1+\varepsilon)\hat{A}_i\right) - \beta D_{KL}(\pi_\theta(\cdot|\bm{s}_t^i) || \pi_{\theta_{\text{old}}}(\cdot|\bm{s}_t^i)) \right]
\end{equation}
and the importance sampling ratio $r_t^i(\theta)$ is defined as:
\begin{equation}
    r_t^i(\theta) = \frac{p_\theta(\bm{z}_{t-1}^i | \bm{z}_t^i, \bm{c})}{p_{\theta_{\text{old}}}(\bm{z}_{t-1}^i | \bm{z}_t^i, \bm{c})}
\end{equation}

\textbf{Stochastic Sampling for Exploration}
Standard Rectified-Flow samplers are deterministic ODEs and therefore lack the stochasticity required for RL exploration. Following prior work~\cite{Maoutsa_2020, song2021scorebasedgenerativemodelingstochastic}, we introduce stochastic transitions using Coefficients-Preserving Sampling (CPS)~\cite{cps_sampling}. The latent update from $t$ to $t-\Delta t$ is
\begin{equation}\label{eq::cps_update}
    \bm{z}_{t-\Delta t} =
    \underbrace{\left((1-(t-\Delta t))\hat{\bm{z}}_0 + (t-\Delta t)\cos\!\left(\tfrac{\eta\pi}{2}\right)\hat{\bm{z}}_1\right)}_{\text{deterministic mean}}
    \;+\;
    \underbrace{(t-\Delta t)\sin\!\left(\tfrac{\eta\pi}{2}\right)\bm{\epsilon}}_{\text{stochastic noise}},
\end{equation}
where $\hat{\bm{z}}_0$ and $\hat{\bm{z}}_1$ are model predictions, $\bm{\epsilon} \sim \mathcal{N}(0, \mathbf{I})$, and $\eta \in [0,1]$ controls noise strength.  
This sampler defines the transition distribution and yields the log-probability needed for the importance ratio computation:
\begin{equation}
    \log p_\theta(\bm{z}_{t-1}^i \mid \bm{z}_t^i, \bm{c}) 
    \propto
    -\left\|\bm{z}_{t-1}^i - \mu_\theta(\bm{z}_t^i, t, \bm{c})\right\|^2,
\end{equation}
where $\mu_\theta$ is the deterministic mean in (Eq.~\ref{eq::cps_update}). Note that for the sampling stage, we did not inject stochasticity in all steps using CPS sampling; instead, we only applied it within a small window of steps, while keeping the rest deterministic, following the Mix-GRPO approach \cite{li2025mixgrpounlockingflowbasedgrpo}.

\subsection{Reward System Design}
The design of the reward function is the key to the post-training stage. Following our multi-agent evaluation system (detailed in \cref{sec::mllm_evaluation}), the reward function $R(\cdot)$ evaluates the generated video $\hat{\bm{v}}$ (decoded from $\bm{z}0$) based on three aspects: lip-sync quality, expressiveness, and motion quality. Each aspect is scored by a MLLM-based evaluation model. For each generated video sample, the final reward is an equally weighted sum of these scores:
\begin{equation}
R_{\text{MLLM}} = \frac{1}{3} \left( R_{\text{lip-sync}}(\hat{\bm{v}}) + R_{\text{expressiveness}}(\hat{\bm{v}}) + R_{\text{motion}}(\hat{\bm{v}}) \right)
\end{equation}
However, using only MLLM-based evaluators leads to a reward hacking phenomenon: the model learns to exploit weaknesses in the evaluators without improving true video fidelity. Empirically, we observe two major artifacts—\emph{jittering} and \emph{color drift}—which MLLMs fail to penalize. To mitigate these issues, we introduce two additional reward components: a frame-level perceptual quality reward based on LPIPS~\cite{zhang2018unreasonableeffectivenessdeepfeatures} , and a motion-smoothness reward based on optical flow consistency using RAFT~\cite{teed2020raftrecurrentallpairsfield}.

\textbf{LPIPS-Based Frame Quality Reward.}
To measure perceptual deviations between generated frames $\hat{\bm{v}}_t$ and reference frames $\bm{v}^{*}_t$, we adopt LPIPS~\cite{zhang2018unreasonableeffectivenessdeepfeatures}. For each image pair, a pretrained network $\mathcal{F}$ extracts channel-normalized features $\hat{y}^{\,l}, \hat{y}^{\,l}_0 \in \mathbb{R}^{H_l \times W_l \times C_l}$ across $L$ layers. LPIPS computes a spatially averaged, channel-weighted distance:
\begin{equation}
    d(\hat{\bm{v}}_t,\bm{v}^{*}_t)
    = \sum_{l} \frac{1}{H_l W_l}
    \sum_{h,w}
    \left\| w^{l} \odot \left( \hat{y}^{\,l}_{hw} - \hat{y}^{\,l}_{0,hw} \right) \right\|_2^2
\end{equation}
where $w^{l} \in \mathbb{R}^{C_l}$ is a learned perceptual weight and $\odot$ denotes channelwise multiplication.  
The frame-quality reward is then defined as the negative temporal average:
\begin{equation}
    R_{\text{perceptual}}
    = - \frac{1}{T} \sum_{t=1}^{T} d(\hat{\bm{v}}_t,\bm{v}^{*}_t)
\end{equation}
This term penalizes perceptual drift, texture degradation, and color inconsistency, complementing MLLM-based semantic signals with a robust low-level fidelity constraint.

\textbf{Optical-Flow Smoothness Reward.}
To penalize temporal jitter, we compute optical flow between consecutive frames using RAFT~\cite{teed2020raftrecurrentallpairsfield}. Let $\mathbf{u}_t \in \mathbb{R}^{2 \times H \times W}$ denote the flow field between $(\hat{\bm{v}}_t, \hat{\bm{v}}_{t+1})$. For $t = 1,\dots,T-2$, we compute the temporal flow change
\begin{equation}
    \Delta \mathbf{u}_t = \mathbf{u}_{t+1} - \mathbf{u}_t
\end{equation}
Following our implementation, we normalize the acceleration by the flow magnitude to obtain a scale-invariant jitter measure:
\begin{equation}
    j_t = \frac{\left\| \Delta \mathbf{u}_t \right\|_2}{\left\| \mathbf{u}_t \right\|_2 + \varepsilon}
\end{equation}
where norms are computed per pixel and averaged spatially. The final consistency reward is defined as the negative temporal average of jitter:
\begin{equation}
    R_{\text{consistency}} = - \frac{1}{T-2} \sum_{t=1}^{T-2} j_t
\end{equation}
such that higher reward corresponds to smoother, more temporally consistent motion.

\textbf{Final Reward Composition}
Because the semantic, perceptual, and motion-smoothness rewards operate on different scales, we first normalize each reward by subtracting its mean and dividing by its standard deviation within the batch. Let $\widetilde{R}_{\text{MLLM}}$, $\widetilde{R}_{\text{perceptual}}$, and $\widetilde{R}_{\text{consistency}}$ denote the normalized rewards. The final reward is then computed as a weighted combination:
\begin{equation}
    R_{\text{final}}
    = \widetilde{R}_{\text{MLLM}}
    + \lambda_1 \, \widetilde{R}_{\text{perceptual}}
    + \lambda_2 \, \widetilde{R}_{\text{consistency}},
\end{equation}
where $\lambda_1$ and $\lambda_2$ control the relative importance of perceptual fidelity and motion smoothness. In our experiments, we set $\lambda_1 = 0.2$ and $\lambda_2 = 0.2$, which provides a stable balance between semantic correctness, visual quality, and temporal consistency.

\section{Experiments}\label{sec::experiments}
\subsection{Setup}
\textbf{SFT Stage}
We adopt a two-stage training pipeline that begins with supervised fine-tuning (SFT) and is later refined with reinforcement learning. In the SFT stage, we initialize our model from BAGEL~\cite{bagel} and train it on a diverse collection of high-quality talking-head video datasets, including several public and internal talking-head video corpora. We perform strict filtering to ensure video quality and speaker consistency, and crop each clip to a 1{:}1 head-centered region (see \cref{app::preprocessing} for details). All videos are standardized to 24~FPS and a maximum spatial resolution of 384~px to match our portrait-animation setup.

For video encoding, we employ the pretrained Wan-VAE~\cite{wan2025wanopenadvancedlargescale} with a patch size of 2, temporal downsampling of 4, and spatial downsampling of 8. Under these settings, a 24~FPS, 384-resolution video produces
\[
(24/4) \times \left(\frac{384}{8 \cdot 2}\right)^2 = 3456 \text{ tokens per second}.
\]
We train the model with a maximum sequence length of 89{,}600 tokens (approximately 25~seconds of video) for 300k steps, using AdamW~\cite{adamw_optimizer} with a learning rate of $5\times10^{-5}$ and $\beta_1=0.9$, $\beta_2=0.95$. During inference, videos are decoded using 25 sampling steps with classifier-free guidance~\cite{ho2022classifierfreediffusionguidance} and a guidance scale of 1.0.

\textbf{RL Post-training Stage}
Building on the strong SFT foundation, we further enhance video quality through reinforcement learning using our Flow-GRPO algorithm. All RL updates are performed on an internal talking-head dataset for 400 optimization steps. During training-time sampling, we follow the CPS procedure described in \cref{sec::approach}, decoding each video in 15 steps. Rather than applying stochastic updates at every step, we select a small window of size $W$ (varied in ablations) in which stochastic updates are applied; the remaining steps are deterministic, and classifier-free guidance is disabled throughout RL sampling.

We optimize using AdamW with a learning rate of $1\times10^{-5}$ and $\beta_1=0.9$, $\beta_2=0.95$, and a batch of 128 generated videos per update. For group-based advantage estimation, each input audio prompt is used to sample $G=4$ candidate videos, and we apply a clipping threshold of $\varepsilon = 0.001$ to stabilize optimization. At inference time, decoding follows the same procedure as in the SFT stage to ensure consistent comparison.

\begin{table*}[t]
\centering
\small
\setlength{\tabcolsep}{8pt}
\renewcommand{\arraystretch}{1.25}
\begin{tabular}{lcccc@{\hspace{3em}}cccc}
\toprule
& \multicolumn{4}{c}{\textbf{In-domain Test Set}}
& \multicolumn{4}{c}{\textbf{Out-domain Test Set}} \\
\cmidrule(lr){2-5} \cmidrule(lr){6-9}
\textbf{Generator}
& \textbf{Lip-sync} & \textbf{Expressive} & \textbf{Motion} & \textbf{Avg}
& \textbf{Lip-sync} & \textbf{Expressive} & \textbf{Motion} & \textbf{Avg} \\
\midrule

\textit{Original}
& 4.03 & 4.33 & 3.43 & 3.93
& 4.74 & 4.42 & 2.93 & 4.03 \\
\midrule

\multicolumn{9}{l}{\textit{Previous Generators}} \\
Echo
& 3.37 & 3.72 & 2.67 & 3.26
& 4.12 & 3.97 & 2.44 & 3.51 \\
Memo
& 3.53 & 3.88 & 2.70 & 3.37
& 4.45 & 4.10 & 2.66 & 3.74 \\
Sonic
& 3.73 & 4.07 & 2.93 & 3.58
& 4.45 & 4.15 & 2.63 & 3.74 \\
\midrule

\multicolumn{9}{l}{\textit{Our Generators}} \\
SFT
& 3.74 & 4.07 & 3.16 & 3.66
& 4.47 & 4.12 & 2.89 & 3.83 \\

\rowcolor{bestrow}
\textbf{RL}
& \textbf{4.42} & \textbf{4.41} & \textbf{3.59} & \textbf{4.14}
& \textbf{4.50} & \textbf{4.17} & \textbf{3.07} & \textbf{3.91} \\

\bottomrule
\end{tabular}
\vspace{-0.5em}
\caption{MLLM-based evaluation of different video generators on in-domain and out-domain test sets.}
\label{tab:generator_comparison}
\vspace{-1em}
\end{table*}

\subsection{Main Results}\label{sec::main_results}

We compare our SFT and RL post-trained models against strong prior baselines, including Echo~\cite{echomimic}, Memo~\cite{memo}, and Sonic~\cite{sonic}. Model quality is evaluated using our MLLM-based automatic assessment framework (MAS-MA) on an in-domain and an out-domain test set. To ensure that our evaluation is not confounded by potential reward hacking introduced during RL post-training, we additionally conduct a human preference study following the same protocol as our automatic evaluation (\cref{sec::human_alignment}). Three independent annotators score each model along lip-sync accuracy, expressiveness, and motion smoothness, and we report the average scores across annotators.

\begin{wraptable}{r}{0.53\textwidth}
\centering
\small
\setlength{\tabcolsep}{6pt}
\renewcommand{\arraystretch}{1.25}
\vspace{-1em}
\begin{tabular}{lcccc}
\toprule
\textbf{Generator} &
\textbf{Lip-sync} & \textbf{Expressive} & \textbf{Motion} & \textbf{Avg} \\
\midrule
Orig  
& 4.78 & 4.76 & 4.68 & 4.74 \\
\midrule
Sonic 
& 3.59 & 3.47 & 3.21 & 3.42 \\

SFT   
& 3.87 & 3.75 & 3.55 & 3.72 \\

\rowcolor{bestrow}
\textbf{RL} 
& \textbf{4.16} & \textbf{4.07} & \textbf{3.85} & \textbf{4.03} \\
\bottomrule
\end{tabular}
\caption{Human annotation results comparing different generators, showing that RL post-training substantially improves generation quality and alignment with human preferences.}
\vspace{-1em}
\label{tab:human_annotation_scores}
\end{wraptable}

\begin{figure}[t]
    \centering
    \small
    \includegraphics[width=0.90\linewidth]{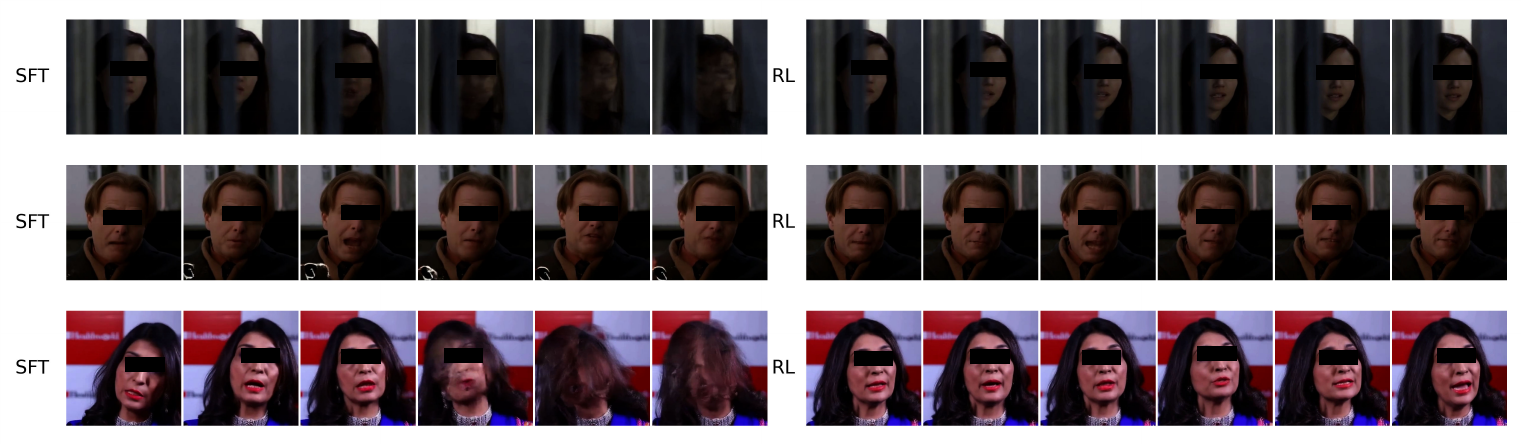}
    \vspace{-1em}
    \caption{Cherrypicked failure cases from our SFT model, including blurred textures and hallucinated artifacts. RL post-training effectively mitigates these issues. The black masks are used to anonymize identities.}
    \label{fig::sft_rl_compare}
\end{figure}

The automatic evaluation results are summarized in \cref{tab:generator_comparison}. Our SFT model already outperforms all prior methods on both the in-domain and out-domain test sets, demonstrating the advantage of initializing portrait animation from a strong MLLM backbone. Building on this baseline, the RL post-trained model further improves performance and achieves the best scores across all evaluation aspects on both test sets. However, because the RL training objective incorporates MLLM-based rewards that overlap with our automatic evaluation, the RL model may benefit from reward hacking effects, which we analyze in \cref{sec::analysis}. We therefore rely on human evaluation as an additional validation of the RL post-training strategy, with results reported in \cref{tab:human_annotation_scores}. As shown, the SFT model is consistently preferred over Sonic in terms of lip-sync, expressiveness, and motion smoothness, though it still trails the original videos. After RL post-training, performance improves across all three dimensions, substantially narrowing the gap to the original videos. Taken together, both automatic and human evaluations demonstrate consistent gains from our training strategies. In \cref{fig::sft_rl_compare}, we show qualitative comparisons between our SFT and RL models, where post-training helps mitigate motion blurring and hallucinated artifacts. More qualitative results are provided in Appendix \ref{app::qualitative_examples}.



\subsection{Analysis and Ablation Studies}\label{sec::analysis}
In previous section, we show the effectiveness of our SFT and RL stages. Here, we provide ablation studies that explored RL post-training design choices and lead to our final reward system design. We explored three settings through ablation studies: (1) noise level $\eta$ in the stochastic update (Eq.~\ref{eq::cps_update}), (2) window size for SDE updates, and (3) reward type and the results are summarized in \cref{tab::ablation-combined}.

\begin{figure}[h!]
    \centering
    \includegraphics[width=0.99\linewidth]{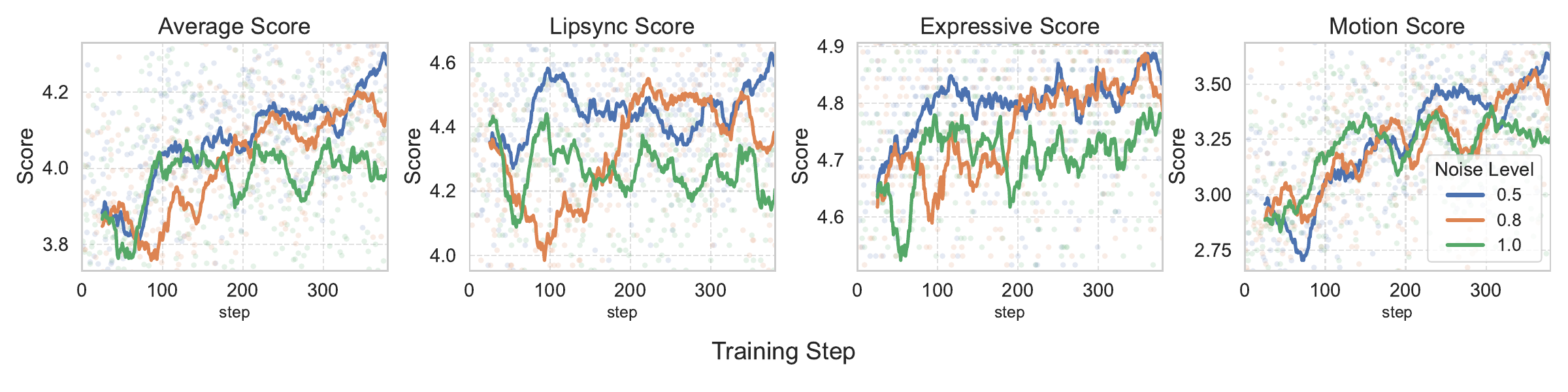}
    \vspace{-2em}
    \caption{Comparison of reward curve during post-training for various noise levels during sampling stage.}
    \label{fig:noise_level_train_log}
\end{figure}

\textbf{Noise Level $\eta$ for Sampling} 
As shown in \cref{tab::ablation-combined} (Ablation~1), a moderate noise level of $\eta=0.5$ during the stochastic update step yields the best performance, providing the largest improvement over the SFT baseline. The reward curves in \cref{fig:noise_level_train_log} further illustrate this trend: high noise ($\eta=1.0$) leads to unstable updates, whereas lower noise ($\eta=0.5$) produces steadily increasing rewards. Qualitatively, models trained with high noise frequently exhibit artifacts such as jittering and temporal color drift (see \cref{fig:visual_bad_case} in the Appendix for examples), while those trained with lower noise show fewer such issues.

\begin{table}[t]
\centering
\small
\setlength{\tabcolsep}{6pt}
\renewcommand{\arraystretch}{1.25}

\begin{tabular}{lcccccc}
\toprule
\textbf{Setting} & \textbf{W} & \textbf{Noise} &
\textbf{Lip-sync} & \textbf{Expressive} & \textbf{Motion} & \textbf{Avg} \\
\midrule

\textbf{SFT Baseline} & N/A & N/A 
& 3.74 & 4.07 & 3.16 & 3.66 \\

\rowcolor{bestrow}
\textbf{RL Baseline ($R_{\text{MLLM}}$)} & \textbf{1} & \textbf{0.5}
& \textbf{4.40} & \textbf{4.40} & \textbf{3.50} & \textbf{4.10} \\

\midrule
\multicolumn{7}{l}{\textbf{Ablation 1: Noise Level}} \\

Noise = 0.8 & 1 & 0.8
& 4.29 (\negval{-0.11}) 
& 4.33 (\negval{-0.07}) 
& 3.48 (\negval{-0.02}) 
& 4.03 (\negval{-0.07}) \\

Noise = 1.0 & 1 & 1.0
& 4.12 (\negval{-0.28}) 
& 4.29 (\negval{-0.11}) 
& 3.41 (\negval{-0.09}) 
& 3.94 (\negval{-0.16}) \\

\midrule
\multicolumn{7}{l}{\textbf{Ablation 2: Window Size}} \\

W = 3 & 3 & 0.5
& 4.22 (\negval{-0.18}) 
& 4.24 (\negval{-0.16}) 
& 3.48 (\negval{-0.02}) 
& 3.98 (\negval{-0.12}) \\

W = 5 & 5 & 0.5
& 4.19 (\negval{-0.21}) 
& 4.28 (\negval{-0.12}) 
& 3.45 (\negval{-0.05}) 
& 3.97 (\negval{-0.13}) \\

\midrule
\multicolumn{7}{l}{\textbf{Ablation 3: Reward Type}} \\

\textbf{$R_{\text{lip-sync}}$} & 1 & 0.5
& 4.36 (\negval{-0.04}) 
& 4.10 (\negval{-0.30}) 
& 3.01 (\negval{-0.49}) 
& 3.83 (\negval{-0.27}) \\

\textbf{$R_{\text{expressiveness}}$} & 1 & 0.5
& 3.96 (\negval{-0.44}) 
& 4.33 (\negval{-0.07}) 
& 2.76 (\negval{-0.74}) 
& 3.68 (\negval{-0.42}) \\

\textbf{$R_{\text{motion}}$} & 1 & 0.5
& 4.13 (\negval{-0.27}) 
& 4.17 (\negval{-0.23}) 
& 3.48 (\negval{-0.02}) 
& 3.92 (\negval{-0.18}) \\

\textbf{$R_{\text{perceptual}} + R_{\text{consistency}}$} & 1 & 0.5
& 4.14 (\negval{-0.26})
& 3.69 (\negval{-0.71})
& 2.06 (\negval{-1.44})
& 3.30 (\negval{-0.80}) \\

\rowcolor{bestrow}
\textbf{$R_{\text{MLLM}} + R_{\text{perceptual}} + R_{\text{consistency}}$} & \textbf{1} & \textbf{0.5}
& \textbf{4.42 (\pos{+0.02})}
& \textbf{4.41 (\pos{+0.01})}
& \textbf{3.59 (\pos{+0.09})}
& \textbf{4.14 (\pos{+0.04})} \\
\bottomrule
\end{tabular}

\caption{All ablations report score differences relative to the RL baseline.
We found that small window size, small noise, and adding perceptual+consistency rewards with MLLM's three aspect rewards produce the best results.}
\label{tab::ablation-combined}
\end{table}

\textbf{Window Size $W$ for Injecting Stochasticity.}
In Ablation~2 of \cref{tab::ablation-combined}, we vary the window size $W$ for stochastic updates while keeping all other factors fixed. Comparing $W \in \{1,3,5\}$, we find that $W=1$ consistently performs best, aligning with prior observations~\cite{longcat_video} that injecting noise at a single step yields more stable optimization. Although the quantitative differences are moderate, qualitative inspection shows that larger windows introduce noticeable temporal jitter and unstable motion. Interestingly, these artifacts often accompany higher motion scores under the MLLM-based evaluator, indicating that the model is exploiting MLLM-based reward system. These findings reveal that MLLM-based reward alone is insufficient to prevent degenerate behaviors, and that stabilizing temporal dynamics requires complementary supervision that is sensitive to low-level artifacts such as jitter.

\textbf{Reward Type.}
Ablation~3 of \cref{tab::ablation-combined} further examines how different reward compositions affect behavior. Optimizing a single aspect reward (Lip-sync, Expressiveness, or Motion) improves that specific metric but degrades the others, leading to characteristic failure modes. Models trained without motion-aware supervision tend to exhibit severe jitter or exaggerated movements, again signaling metric-driven reward hacking.

\begin{figure}[h!]
    \centering
    \includegraphics[width=0.95\linewidth]{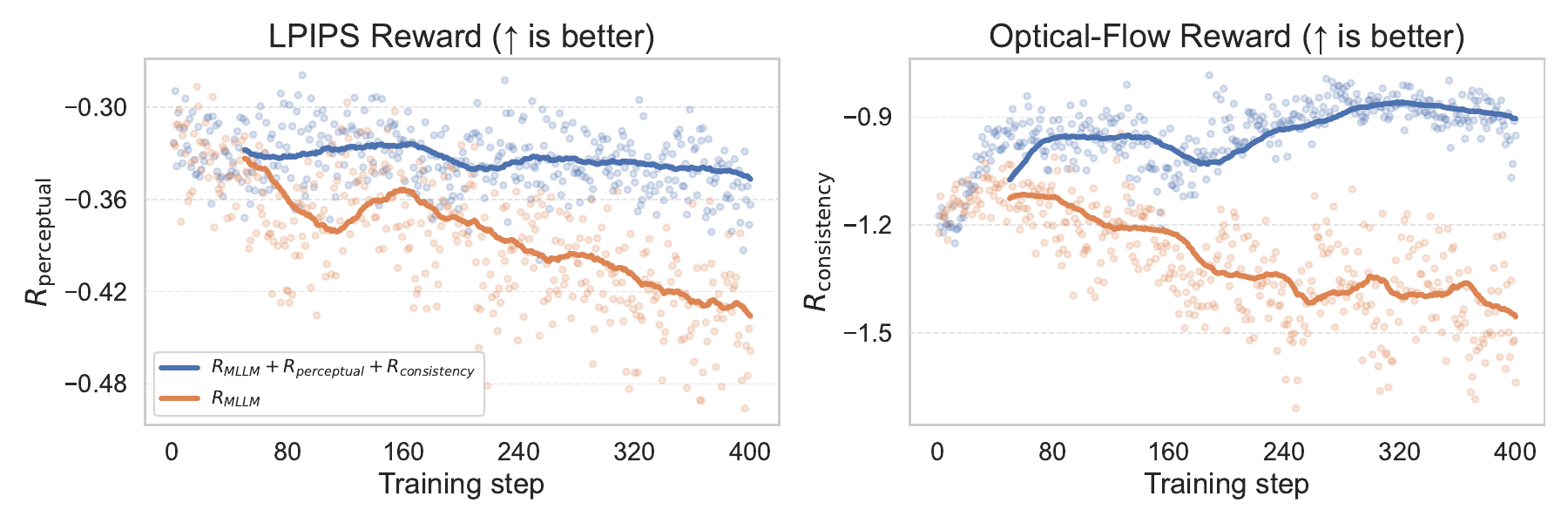}
    \vspace{-1em}
    \caption{
    Comparing training-time metrics—$R_{\text{perceptual}}$ and $R_{\text{consistency}}$—reveals a clear gap between using only the MLLM-based reward and augmenting it with perceptual and consistency terms. Without these additional rewards, both metrics deteriorate sharply, aligning with the jittering and color drift observed in our qualitative results.
    }
    \label{fig::train_reg_curve}
\end{figure}

To counteract these behaviors, we augment the MLLM-based reward with two additional components: a perceptual term and an optical-flow jitter penalty that explicitly discourages unstable motion. This combined reward yields modest improvements in MLLM evaluation metrics but, more importantly, produces qualitatively more natural and stable videos, with no observable jittering or color drift (examples in Appendix \ref{app::qualitative_examples}) and supported by human evaluation in \cref{sec::main_results}. Moreover, as shown in \cref{fig::train_reg_curve}, training without these additional terms causes LPIPS and Optical Flow Consistency scores to deteriorate, confirming that the model overfits to $R_{\text{MLLM}}$ and begins exploiting its blind spots.

Finally, using only the perceptual and consistency components—without the MLLM reward—induces a different form of reward hacking: the model collapses toward nearly static outputs that maximize temporal smoothness but perform poorly on all other aspects, yielding a substantially lower average score (3.3 versus 4.10 for the SFT baseline). These observations underscore the importance of combining high-level MLLM judgments with low-level perceptual and flow-based constraints to achieve stable, robust, and human-aligned improvements in generation quality.

\textbf{Summary}
Our findings indicate that the optimal configuration is moderate noise, a minimal stochastic window, and a combined multi-aspect reward. To further mitigate reward hacking in the MLLM-based objective, we add perceptual and consistency rewards, which effectively regularize training and produce more natural, artifact-free videos. We adopt this setup in our main RL post-training experiments shown in \cref{sec::main_results}.

\section{Related Work}

\subsection{Evaluation Metrics for Portrait Animation}
Prior work \cite{sonic, memo, echomimic} typically evaluates avatar generation using distributional metrics such as FID \cite{fid} and FVD \cite{fvd}, SyncNet-based LSE-C/D \cite{syncnet} for lip-sync accuracy, and pixel-level measures including PSNR and SSIM. While widely adopted, these metrics often correlate poorly with human perception, as they emphasize low-level visual similarity and fail to reflect higher-level attributes such as expressiveness and motion naturalness.

More recently, \citep{quignon2025thevalevaluationframeworktalking} proposed a multi-aspect evaluation framework for portrait animation that decomposes quality into three dimensions—Quality, Naturalness, and Synchronization—and introduces corresponding automatic metrics. In contrast, \citep{ding2025klingavatargroundingmultimodalinstructions} evaluate avatar generation using higher-level criteria such as lip-sync, expressiveness, instruction controllability, and identity preservation, but rely exclusively on human Good/Same/Bad (GSB) judgments rather than automatic evaluation. Similarly, \citep{huang2025liveavatarstreamingrealtime} assess naturalness, synchronization, and consistency using human ratings on a 1–100 scale with 20 annotators. Our human annotation protocol is partially inspired by these contemporaneous efforts; however, to the best of our knowledge, we are the first to employ an MLLM-based system for both evaluation and reinforcement-based post-training of portrait animation models. 

\subsection{Reinforcement Learning for Video Generation}
Reinforcement learning (RL) has recently gained traction in image and video generation \cite{flow_grpo, tempflow_grpo, li2025mixgrpounlockingflowbasedgrpo, longcat_video, chen2025blip3onextfrontiernativeimage}, enabling models to directly optimize complex, task-specific reward objectives. Advances on the algorithmic side—such as improved sampling schemes and more stable optimization strategies—have further enhanced training efficiency and robustness \cite{cps_sampling, tempflow_grpo}. Despite this progress, most prior work relies on reward signals derived from HPS \cite{wu2023humanpreferencescorev2} or CLIP-Score \cite{hessel2022clipscorereferencefreeevaluationmetric}, both of which are vulnerable to reward hacking during post-training. Recent efforts, such as GRPO-Guard \cite{wang2025grpoguardmitigatingimplicitoveroptimization}, propose to address this issue by regulating clipping to prevent over-optimization. Developing fine-grained, reliable, and robust reward systems for video generation remains an open research challenge.

\section{Conclusion}
We present \method, a reinforcement-learning–enhanced framework for high-quality portrait animation using an \arflow backbone. By combining a multimodal pretrained model with a composite reward system—integrating MLLM-based multi-aspect evaluations with pixel-level perceptual terms and an optical-flow jitter penalty—we achieve substantial improvements in lip-sync accuracy, facial expressiveness, and motion naturalness. Our results highlight the importance of fine-grained, human-aligned reward design for advancing realistic audio-driven portrait animation.


\bibliographystyle{unsrt}
\bibliography{main}

\clearpage

\beginappendix

\section{Human Annotation}\label{app::human_annotation}

After collecting videos from different generators, we recruit three human annotators to evaluate each clip along three dimensions: lip-sync quality, facial expressiveness, and motion smoothness (naturalness). Annotators are provided with detailed scoring rubrics (shown in \cref{tab:lipsync_eval}, \cref{tab:expression_eval}, and \cref{tab:motion_eval}) and are instructed to watch each video in its entirety before assigning a score from 1 (worst) to 5 (best) for each aspect.

In addition to aspect-level scoring, annotators are asked to provide an overall ranking of the videos within each set. For example, given an original video (A) and generated videos from Sonic \cite{sonic} (B), Memo \cite{memo} (C), and Echomimics \cite{echomimic} (D), annotators rank the four videos from best to worst (e.g., $C > A \geq B = D$). Ties and partial orderings are allowed to account for videos of comparable quality.

To compute the agreement between automatic evaluation methods and human judgments (\cref{sec::human_alignment}), we retain only video pairs for which all three annotators agree on the relative ordering. Pairs with disagreement (e.g., two annotators prefer A over B while one prefers B over A) are excluded. This filtering results in approximately 80\% of all annotated pairs being used for match-rate computation.

\begin{table}[h!]
\centering
\small
\begin{tabular}{c p{1.5cm} p{12cm}}
\hline
\textbf{Score} & \textbf{Guideline} & \textbf{Detailed Criteria (Precision, Timing, \& Naturalness)} \\
\hline
5 & Excellent &
\textbf{Precision:} Mouth shapes perfectly match all spoken phonemes. 
\textbf{Timing:} Lip movements are perfectly synchronized with the audio, with no noticeable delay or lead.
\textbf{Naturalness:} Movements are fluid and human-like, with no warping or visual artifacts. \\

4 & Good &
\textbf{Precision:} Mouth shapes are correct for the vast majority of sounds, with only subtle inaccuracies.
\textbf{Timing:} Synchronization is excellent, with only an almost imperceptible lag or lead.
\textbf{Naturalness:} Movements look natural but may lack fine-grained nuances. \\

3 & Fair &
\textbf{Precision:} Mouth shapes are generally correct but imprecise or simplified.
\textbf{Timing:} A slight but consistent delay or lead is noticeable.
\textbf{Naturalness:} Movements appear somewhat robotic or overly smooth; minor blurring artifacts may be present. \\

2 & Poor &
\textbf{Precision:} Mouth shapes are frequently incorrect.
\textbf{Timing:} Audio–visual synchronization is clearly off and distracting.
\textbf{Naturalness:} Movements are jerky or exhibit unnatural “flapping,” with obvious warping. \\

1 & Bad &
\textbf{Precision:} Lip movements have little to no correlation with speech.
\textbf{Timing:} Audio and lip movements are severely out of sync.
\textbf{Naturalness:} Mouth motion is highly distorted, nonsensical, or completely static. \\

\hline
\end{tabular}
\caption{Human evaluation rubric for lip-sync quality (1--5).}
\label{tab:lipsync_eval}
\end{table}

\begin{table}[h!]
\centering
\small
\begin{tabular}{c p{1.5cm} p{12cm}}
\hline
\textbf{Score} & \textbf{Guideline} & \textbf{Detailed Criteria (Emotional Match \& Realism)} \\
\hline
5 & Excellent &
\textbf{Emotional Match:} Facial expressions perfectly reflect the emotional tone of the audio.
\textbf{Realism:} Includes natural micro-expressions and appears fully human and genuine. \\

4 & Good &
\textbf{Emotional Match:} Expressions largely match the audio’s emotion, with minor inconsistencies.
\textbf{Realism:} Looks realistic but may feel slightly acted or less nuanced. \\

3 & Fair &
\textbf{Emotional Match:} Emotion is generally appropriate but muted, delayed, or underspecified.
\textbf{Realism:} The face appears mildly uncanny or artificial. \\

2 & Poor &
\textbf{Emotional Match:} Expression mismatches the audio emotion or remains stuck in a neutral state.
\textbf{Realism:} Expressions look highly unnatural and artificial. \\

1 & Bad &
\textbf{Emotional Match:} Expression is completely incorrect, nonsensical, or emotionless.
\textbf{Realism:} Face resembles a static mask or mannequin. \\

\hline
\end{tabular}
\caption{Human evaluation rubric for facial expressiveness (1--5).}
\label{tab:expression_eval}
\end{table}

\begin{table}[h!]
\centering
\small
\begin{tabular}{c p{1.5cm} p{12cm}}
\hline
\textbf{Score} & \textbf{Guideline} & \textbf{Detailed Criteria (Head Movement, Subtle Motion \& Consistency)} \\
\hline
5 & Excellent &
\textbf{Head Movement:} Smooth, fluid, and purposeful.
\textbf{Subtle Motion:} Natural blinks, head tilts, and minor posture shifts are present.
\textbf{Consistency:} Motion is coherent and human-like throughout. \\

4 & Good &
\textbf{Head Movement:} Smooth and stable, but somewhat simple or repetitive.
\textbf{Subtle Motion:} Blinking appears natural; other subtle motions may be limited.
\textbf{Consistency:} Motion remains plausible with no major artifacts. \\

3 & Fair &
\textbf{Head Movement:} Slightly stiff, jerky, or floaty.
\textbf{Subtle Motion:} Blinking is unnatural in cadence or absent for periods.
\textbf{Consistency:} Occasional robotic or locked head poses occur. \\

2 & Poor &
\textbf{Head Movement:} Robotic, jittery, or laggy.
\textbf{Subtle Motion:} No natural blinks or fine motions are observed.
\textbf{Consistency:} Motion is highly unnatural and distracting. \\

1 & Bad &
\textbf{Head Movement:} Head is completely static.
\textbf{Subtle Motion:} No blinking or facial motion at all.
\textbf{Consistency:} The lack of motion is constant and clearly artificial. \\

\hline
\end{tabular}
\caption{Human evaluation rubric for video smoothness and motion quality (1--5).}
\label{tab:motion_eval}
\end{table}

\newpage
\section{Preprocessing Pipeline}\label{app::preprocessing}
To construct training data for portrait animation, we apply a multi-stage preprocessing pipeline designed to ensure high-quality and well-aligned audio-visual content:
\begin{itemize}[leftmargin=2em, itemsep=2pt, topsep=2pt]
    \item \textbf{Speech-Based Filtering}: We first perform automatic speech recognition (ASR) and speaker diarization on all videos using in-house models. Clips containing more than 10\% overlapping speech from multiple speakers are removed, as such cases often exhibit audio-visual misalignment. We also discard clips identified as non-English based on their transcriptions.
    \item \textbf{Visual-Based Filtering}: We apply light-ASD \cite{Liao_2023_CVPR} for active speaker detection and remove frames whose final active-speaker confidence is below 0.5. For datasets that provide face-direction annotations, we further filter out clips that are not labeled as having either a \texttt{frontal} or \texttt{side} face orientation.
    \item \textbf{Face Detection and Cropping}: We detect faces in each frame using the \texttt{buffalo\_sc} \cite{Deng_2019_CVPR} from InsightFace. If no face is detected for more than one second (corresponding to 24 consecutive frames at 24 FPS), the video is split into separate sub-clips. For each detected face, we crop a square (1:1 aspect ratio) region centered on the face and enlarge the bounding box by a factor of 1.5. For example, an original bounding box of size $(224,168)$ is first converted to $(224,224)$ and then expanded to $(336,336)$. In cases where the enlarged box exceeds frame boundaries, we adjust its size or position to ensure a valid crop.
\end{itemize}

\section{Evaluation Prompts}\label{app::evaluation_prompts}
To automatically evaluate generated portrait animation, we design separate prompts that explicitly control model behavior by restricting attention to a single evaluation aspect at a time. The detailed prompts for assessing lip-sync quality, facial expressiveness, and motion smoothness are presented in \cref{tab:prompt-lipsync}, \cref{tab:prompt-expressive}, and \cref{tab:prompt-motion}, respectively.

%

\begin{table*}[h!]
\centering
\small
\setlength{\tabcolsep}{10pt}
\renewcommand{\arraystretch}{1.15}

\begin{tabularx}{\textwidth}{@{}>{\bfseries}p{0.1\textwidth}X@{}}
\toprule
Prompt Name & Lip-Sync Evaluation Prompt (Single-Video, Single-Aspect) \\
\midrule

Role &
You are a specialized analyst focusing on audio-visual synchronization in digital media.
Your expertise lies in phonetics and the precise mapping of mouth movements to speech. \\
\addlinespace[2pt]

Context &
You will analyze a single video. Your \emph{sole focus} is to evaluate and score its lip-sync quality.
You must ignore all other aspects like emotional expression or head movement. \\
\addlinespace[2pt]

Primary Objective &
Conduct a rigorous analysis and scoring of the lip-sync quality for this video.
Your analysis must justify the score given and return a JSON dictionary with the score. \\
\addlinespace[2pt]

Evaluation Criteria &
\begin{itemize}[nosep]
  \item \textbf{Precision:} How accurately do the lip movements match vowel/consonant shapes (visemes)?
  \item \textbf{Timing:} Any noticeable delay/lead between audio and corresponding lip movements?
  \item \textbf{Naturalness:} Human/fluid vs.\ artificial/robotic/flappy; warping or artifacts around the mouth?
\end{itemize} \\
\addlinespace[2pt]

Scoring Rubric (1--5) &
\begin{itemize}[nosep]
  \item \textbf{5 (Perfect):} Precision matches all phonemes; timing perfectly synchronized; natural, fluid, no artifacts.
  \item \textbf{4 (Excellent):} Mostly correct visemes with subtle inaccuracies; near-imperceptible lag/lead; slightly scripted.
  \item \textbf{3 (Good/Acceptable):} Generally correct but simplified/puppet-like; slight consistent lag/lead; minor artifacts possible.
  \item \textbf{2 (Poor):} Frequently incorrect visemes; clearly off-sync and distracting; robotic/jerky, obvious warping/flapping.
  \item \textbf{1 (Failure):} No correlation to speech; severely out of sync; bizarre/distorted/nonsensical or static mouth.
\end{itemize} \\
\addlinespace[2pt]

Output Constraints &
\begin{enumerate}[nosep]
  \item Put all reasoning and analysis \textbf{inside} \texttt{<think>...</think>}.
  \item After \texttt{</think>}, output \textbf{only} a valid JSON object with key \texttt{"score"} and an integer \texttt{1--5}.
  \item Do \textbf{not} include natural language, commentary, markdown, or extra quotes outside the \texttt{<think>} block.
  \item The final JSON must be valid and contain \textbf{only} the score.
\end{enumerate} \\
\addlinespace[2pt]

Example Output &
\vspace{-5pt}
\begin{tcolorbox}[
  enhanced,
  colback=gray!4,
  colframe=gray!35,
  boxrule=0.6pt,
  arc=2mm,
  left=1.5mm, right=1.5mm, top=1.0mm, bottom=1.0mm
]
\ttfamily\footnotesize
\textless think\textgreater\\
(detailed reasoning here, free-form, any text allowed)\\
\textless /think\textgreater\\[2pt]
\{''score'': integer between 1 and 5\}
\end{tcolorbox}
\\
\bottomrule
\end{tabularx}

\caption{Prompt used for lip-sync evaluation. The model is instructed to assess only audio-visual synchronization and to return a single 1--5 score in JSON format.}
\label{tab:prompt-lipsync}
\end{table*}


\begin{table*}[h!]
\centering
\small
\setlength{\tabcolsep}{10pt}
\renewcommand{\arraystretch}{1.15}

\begin{tabularx}{\textwidth}{@{}>{\bfseries}p{0.1\textwidth}X@{}}
\toprule
Prompt Name & Facial Expressiveness Evaluation Prompt (Single-Video, Single-Aspect) \\
\midrule

Role &
You are a specialized analyst in human emotion and facial expression, with expertise in facial action coding systems (FACS) and psychological realism in digital actors. \\
\addlinespace[2pt]

Context &
You will analyze a single video. Your \emph{sole focus} is to evaluate and score its facial expressiveness and emotional authenticity.
You must apply the scoring rubric below to assign a score from 1 to 5. \\
\addlinespace[2pt]

Primary Objective &
Conduct a rigorous analysis of the video's facial expressiveness, assign a score based on the provided rubric, and return a JSON dictionary with the score. \\
\addlinespace[2pt]

Evaluation Criteria &
\begin{itemize}[nosep]
  \item \textbf{Emotional Match:} Does the facial expression accurately reflect the audio’s tone, sentiment, and emphasis?
  \item \textbf{Realism \& Subtlety:} Are there subtle micro-expressions (e.g., brow movements, eye crinkling, cheek twitches), or does the face appear uncanny or plastic?
  \item \textbf{Natural Dynamics:} Do expressions transition smoothly with speech, or appear robotic, delayed, or disconnected?
\end{itemize} \\
\addlinespace[2pt]

Scoring Rubric (1--5) &
\begin{itemize}[nosep]
  \item \textbf{5 (Excellent):} Perfect emotional alignment with the audio; rich micro-expressions; fully genuine and human-like.
  \item \textbf{4 (Good):} Strong emotional match with minor inconsistencies; realistic but slightly acted or less nuanced.
  \item \textbf{3 (Average):} Generally appropriate emotion but muted, delayed, or underspecified; mildly uncanny or plastic.
  \item \textbf{2 (Poor):} Clear emotional mismatch or fixed neutral/awkward state; highly unnatural and artificial.
  \item \textbf{1 (Very Poor):} Completely incorrect or absent emotion; static, mask-like, or mannequin-like face.
\end{itemize} \\
\addlinespace[2pt]

Output Constraints &
\begin{enumerate}[nosep]
  \item Put all reasoning and analysis \textbf{inside} \texttt{<think>...</think>}.
  \item After \texttt{</think>}, output \textbf{only} a valid JSON object with key \texttt{"score"} and an integer \texttt{1--5}.
  \item Do \textbf{not} include natural language, commentary, markdown, or extra quotes outside the \texttt{<think>} block.
  \item The final JSON must be valid and contain \textbf{only} the score.
\end{enumerate} \\
\addlinespace[2pt]

Example Output &
\vspace{-5pt}
\begin{tcolorbox}[
  enhanced,
  colback=gray!4,
  colframe=gray!35,
  boxrule=0.6pt,
  arc=2mm,
  left=1.5mm, right=1.5mm, top=1.0mm, bottom=1.0mm
]
\ttfamily\footnotesize
\textless think\textgreater\\
(detailed reasoning here, free-form, any text allowed)\\
\textless /think\textgreater\\[2pt]
\{''score'': integer between 1 and 5\}
\end{tcolorbox}
\\

\bottomrule
\end{tabularx}

\caption{Prompt used for evaluating facial expressiveness and emotional authenticity. The model assesses emotional alignment and realism of facial expressions and returns a single 1--5 score in JSON format.}
\label{tab:prompt-expressive}
\end{table*}


\begin{table*}[h!]
\centering
\small
\setlength{\tabcolsep}{10pt}
\renewcommand{\arraystretch}{1.15}

\begin{tabularx}{\textwidth}{@{}>{\bfseries}p{0.1\textwidth}X@{}}
\toprule
Prompt Name & Motion Smoothness and Background Quality Evaluation Prompt (Single-Video, Single-Aspect) \\
\midrule

Role &
You are a specialized analyst in motion quality and video composition.
Your expertise lies in identifying motion artifacts, assessing movement naturalness, and spotting background anomalies in synthesized media. \\
\addlinespace[2pt]

Context &
You will analyze a single video. Your \emph{sole focus} is to evaluate and score its motion smoothness and background integrity.
You must ignore all other aspects such as lip-sync or facial emotion. \\
\addlinespace[2pt]

Primary Objective &
Conduct a rigorous analysis of the video’s motion quality and background stability, assign a score using the provided rubric, and return a JSON dictionary with the score. \\
\addlinespace[2pt]

Evaluation Criteria &
\begin{itemize}[nosep]
  \item \textbf{Head Movement:} Are head motions smooth, fluid, and purposeful, or stiff, jittery, and robotic?
  \item \textbf{Subtle Motions:} Are natural micro-movements (e.g., blinking, slight tilts, posture shifts) present and realistic?
  \item \textbf{Consistency:} Does motion remain coherent and human-like throughout, without locking or abrupt artifacts?
  \item \textbf{Background Quality:} Is the background stable, with clean edges around hair and shoulders, or does it exhibit warping, shimmering, or distortion?
\end{itemize} \\
\addlinespace[2pt]

Scoring Rubric (1--5) &
\begin{itemize}[nosep]
  \item \textbf{5 (Excellent):} Perfectly smooth and purposeful head motion; rich subtle movements; fully consistent and human-like; background entirely stable with no artifacts.
  \item \textbf{4 (Good):} Smooth and consistent motion but somewhat simple or repetitive; natural blinking; minor, non-distracting background artifacts.
  \item \textbf{3 (Average):} Slightly stiff, jerky, or floaty motion; unnatural blinking cadence or frequent stillness; noticeable but moderate background warping or blur.
  \item \textbf{2 (Poor):} Clearly robotic or jittery motion with unnatural jerks; no subtle movements; distracting instability; obvious background distortion or melting.
  \item \textbf{1 (Failure):} Completely static or locked head; frozen face with no movement; entirely artificial appearance; severe background and subject distortion.
\end{itemize} \\
\addlinespace[2pt]

Output Constraints &
\begin{enumerate}[nosep]
  \item Put all reasoning and analysis \textbf{inside} \texttt{<think>...</think>}.
  \item After \texttt{</think>}, output \textbf{only} a valid JSON object with key \texttt{"score"} and an integer \texttt{1--5}.
  \item Do \textbf{not} include natural language, commentary, markdown, or extra quotes outside the \texttt{<think>} block.
  \item The final JSON must be valid and contain \textbf{only} the score.
\end{enumerate} \\
\addlinespace[2pt]

Example Output &
\vspace{0pt}
\begin{tcolorbox}[
  enhanced,
  colback=gray!4,
  colframe=gray!35,
  boxrule=0.6pt,
  arc=2mm,
  left=1.5mm, right=1.5mm, top=1.0mm, bottom=1.0mm
]
\ttfamily\footnotesize
\textless think\textgreater\\
(detailed reasoning here, free-form, any text allowed)\\
\textless /think\textgreater\\[2pt]
\{''score'': integer between 1 and 5\}
\end{tcolorbox}
\\

\bottomrule
\end{tabularx}

\caption{Prompt used for evaluating motion smoothness and background integrity. The model assesses head motion dynamics, subtle movements, and background stability, and returns a single 1--5 score in JSON format.}
\label{tab:prompt-motion}
\end{table*}

\section{Qualitative Examples}\label{app::qualitative_examples}

We provide additional qualitative examples to complement the quantitative
results presented in the main paper. Identities are anonymized with black masks.

\begin{figure}[h!]
    \centering
    \small
    \includegraphics[width=0.95\linewidth]{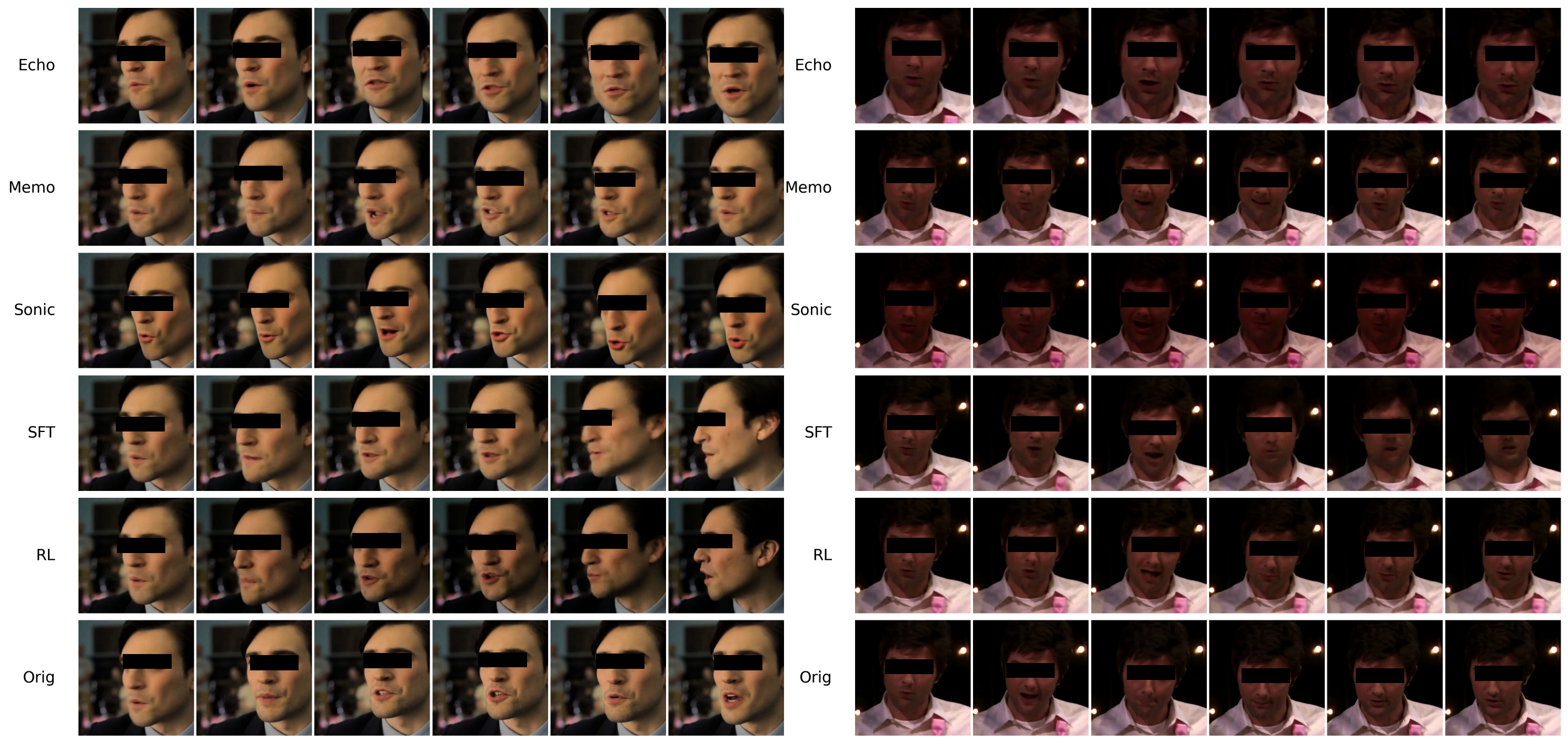}
    \vspace{-1em}
    \caption{Qualitative examples comparing prior methods with our SFT and RL
    post-trained models. Our approach generates videos with more natural
    movement, better lip-sync, and improved expressiveness. Black masks are
    used to anonymize identities.}
    \label{fig::teaser_figure}
\end{figure}

\begin{figure}[h!]
    \centering
    \includegraphics[width=0.95\linewidth]{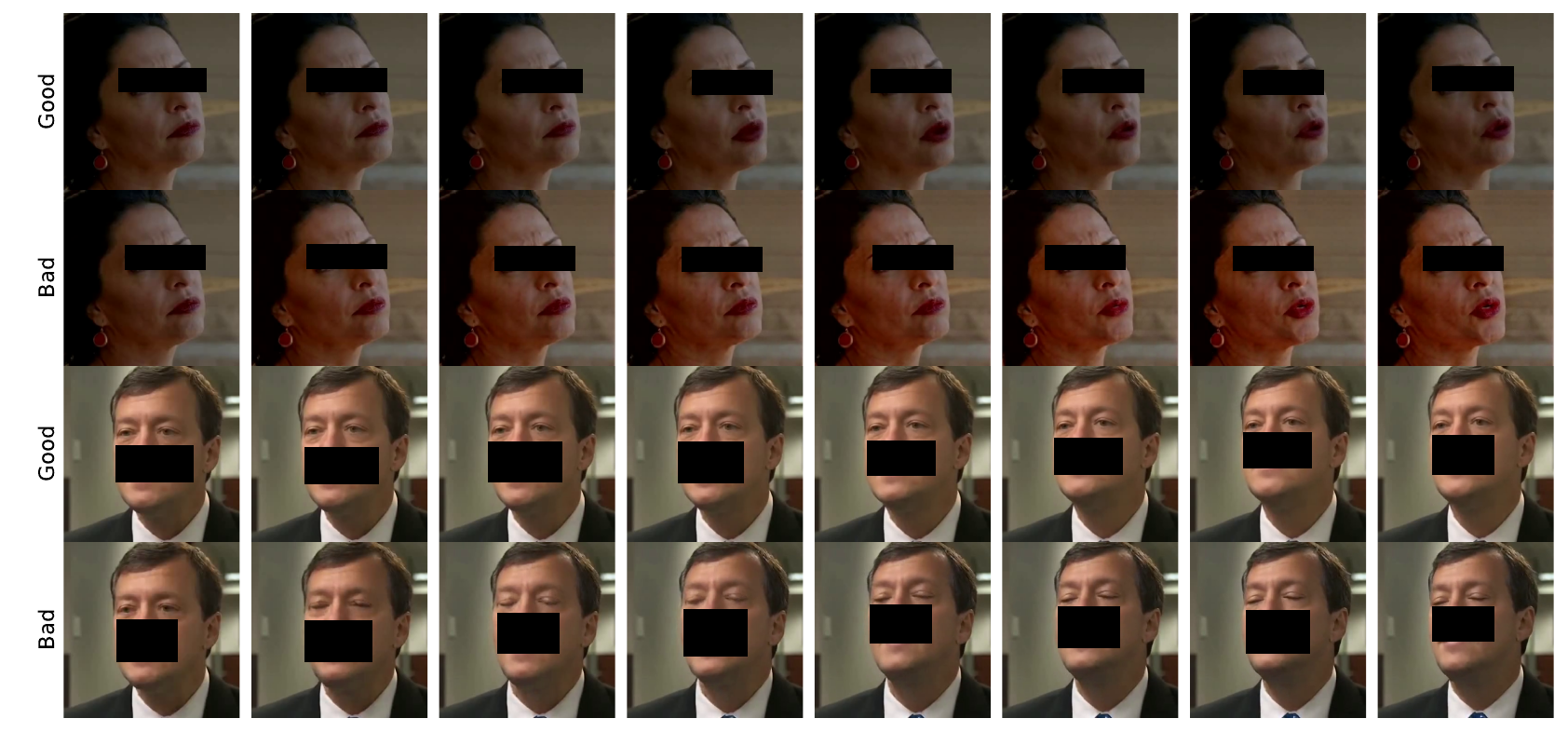}
    \vspace{-1em}
    \caption{Examples of generated video frames illustrating the effect of
    different post-training configurations. \textbf{Good cases (Rows 1 and 3):}
    Our final post-trained model (all three aspect rewards with perceptual and
    consistency rewards, $W{=}1$, $\eta{=}0.5$) produces consistent identity,
    stable colors, and smooth temporal dynamics.
    \textbf{Bad case 1 (Row 2):} Training with a large stochastic window leads
    to strong temporal color drift and bad texture.
    \textbf{Bad case 2 (Row 4):} Training with only the lip-sync reward
    introduces jittering and unnatural motion, including exaggerated eye
    movements. Black masks are used to anonymize identities.}
    \label{fig:visual_bad_case}
\end{figure}

\cref{fig::teaser_figure} compares generated frames from prior methods and our
models. \cref{fig:visual_bad_case} contrasts good and bad cases under different
post-training configurations: well-tuned settings yield stable identity and
smooth motion, whereas suboptimal choices lead to temporal color drift
(large stochastic window) or jittering and exaggerated motion (lip-sync-only
reward).

\end{document}